\documentclass{article}

\usepackage[final]{neurips_2025}

\usepackage{paralist}
\usepackage{microtype}
\usepackage{graphicx}
\usepackage{subfigure}
\usepackage{booktabs} 

\usepackage{bbm}

\usepackage{wrapfig}
\usepackage{hyperref}
\usepackage{amsmath}
\usepackage{amssymb}
\usepackage{mathtools}
\usepackage{amsthm}
\usepackage{multirow}
\usepackage{multicol}
\usepackage[utf8]{inputenc}
\usepackage{tcolorbox}
\usepackage{float}

\usepackage[capitalize,noabbrev]{cleveref}

\theoremstyle{plain}

\theoremstyle{definition}

\theoremstyle{remark}

\usepackage{colortbl}

\newcommand{\method}{\textsc{PPG-Distill}}
\newcommand{\vanilla}{Global \textit{KD}}

\title{\method{}: Efficient Photoplethysmography Signals Analysis via Foundation Model Distillation}

\author{%
Juntong Ni\textsuperscript{1},
\enskip Saurabh Kataria\textsuperscript{2},
\enskip Shengpu Tang\textsuperscript{1},
\enskip Carl Yang\textsuperscript{1},
\enskip Xiao Hu\textsuperscript{2},
\enskip Wei Jin\textsuperscript{1}\\
\textsuperscript{1}Department of Computer Science, Emory University \\
\textsuperscript{2}Nell Hodgson Woodruff School of Nursing, Emory University \\
\texttt{\{firstname.lastname\}@emory.edu}
}

\begin{document}

\maketitle

\begin{abstract}
Photoplethysmography (PPG) is widely used in wearable health monitoring, yet large PPG foundation models remain difficult to deploy on resource-limited devices. We present \textbf{\method{}}, a knowledge distillation framework that transfers both global and local knowledge through prediction-, feature-, and patch-level distillation. \method{} incorporates \textit{morphology distillation} to preserve local waveform patterns and \textit{rhythm distillation} to capture inter-patch temporal structures. On heart rate estimation and atrial fibrillation detection, \method{} improves student performance by up to $21.8\%$ while achieving $7\times$ faster inference and reducing memory usage by $19\times$, enabling efficient PPG analysis on wearables. Our code is available at \url{https://github.com/LingFengGold/PPG-Distill}.

\end{abstract}

\section{Introduction}
\label{sec:introduction}



Wearable sensors that are unobtrusive, widely accessible, and cost-effective have demonstrated strong potential for real-time health monitoring. Among these, photoplethysmography (PPG), an inherently time-series signal that captures continuous variations in blood volume over time, has become a widely used modality in smartwatches~\cite{chen2025gpt, saha2025pulse}. Its popularity arises from enabling non-invasive physiological assessment without requiring firm skin attachment~\cite{saha2025pulse, shi2009photoplethysmography}. The rich information in PPG arises from its local waveform morphology, which reflects cardiovascular events, and its long-range structural rhythm, reflecting periodicity and autonomic regulation.
These properties enable applications from cardiovascular monitoring~\cite{reiss2019deep, schmidt2018introducing,sarhaddi2022comprehensive,wang2023pulsedb, he2022new,ali2024comparison,pimentel2016toward}, clinical diagnostics~\cite{torres2020multi, poh2018diagnostic,clifford2015physionet,lazazzera2020detection,xu2025ecg, liu2025graph}, to mental state assessment~\cite{zhu2023stress,kontaxis2020photoplethysmographic,wang2024classifying}.

\begin{figure*}[t!]
    \centering
    \begin{minipage}[t]{0.69\textwidth}
        \centering
        \includegraphics[width=\textwidth]{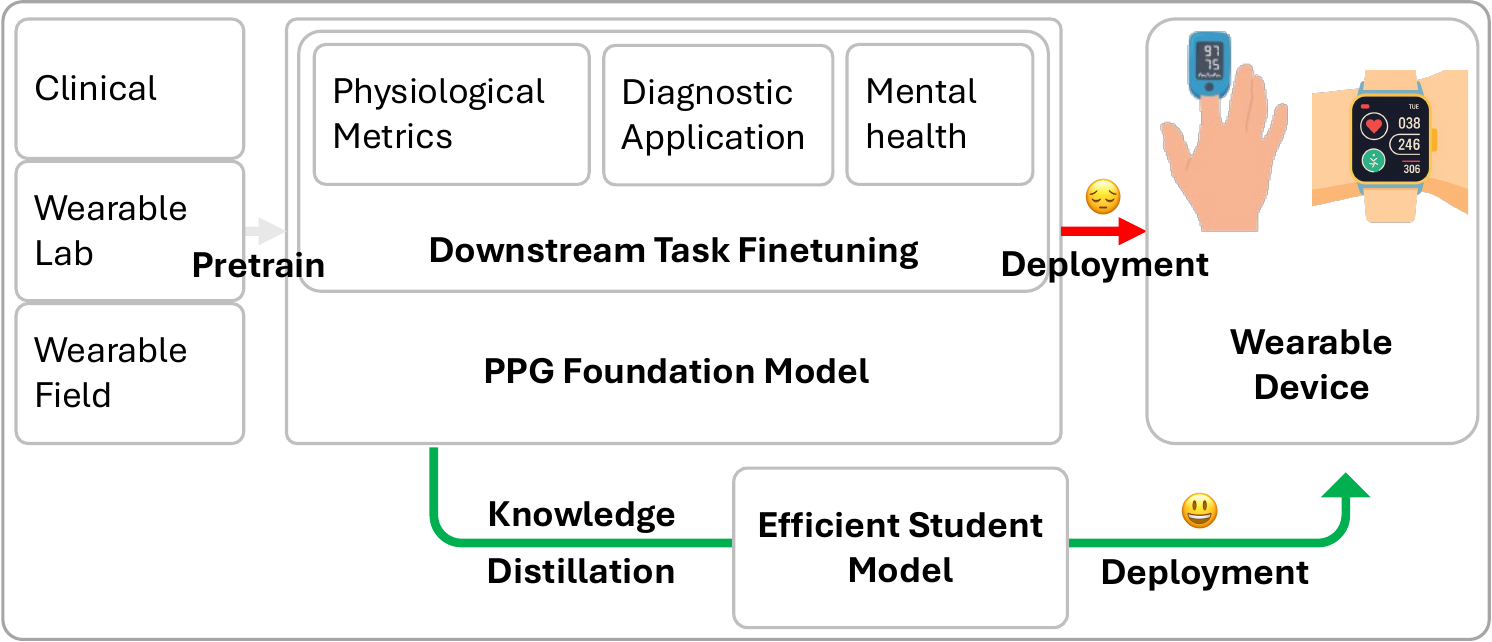}
        \vspace{-1em}
        \caption{Illustration of our motivation. PPG foundation models are pretrained and finetuned for downstream tasks, but direct deployment on wearables is costly. \textit{KD} produces efficient student models suitable for wearable deployment.}
        \label{fig:motivation}
    \end{minipage}%
    \hfill
    \begin{minipage}[t]{0.29\textwidth}
        \centering
        \includegraphics[width=\textwidth]{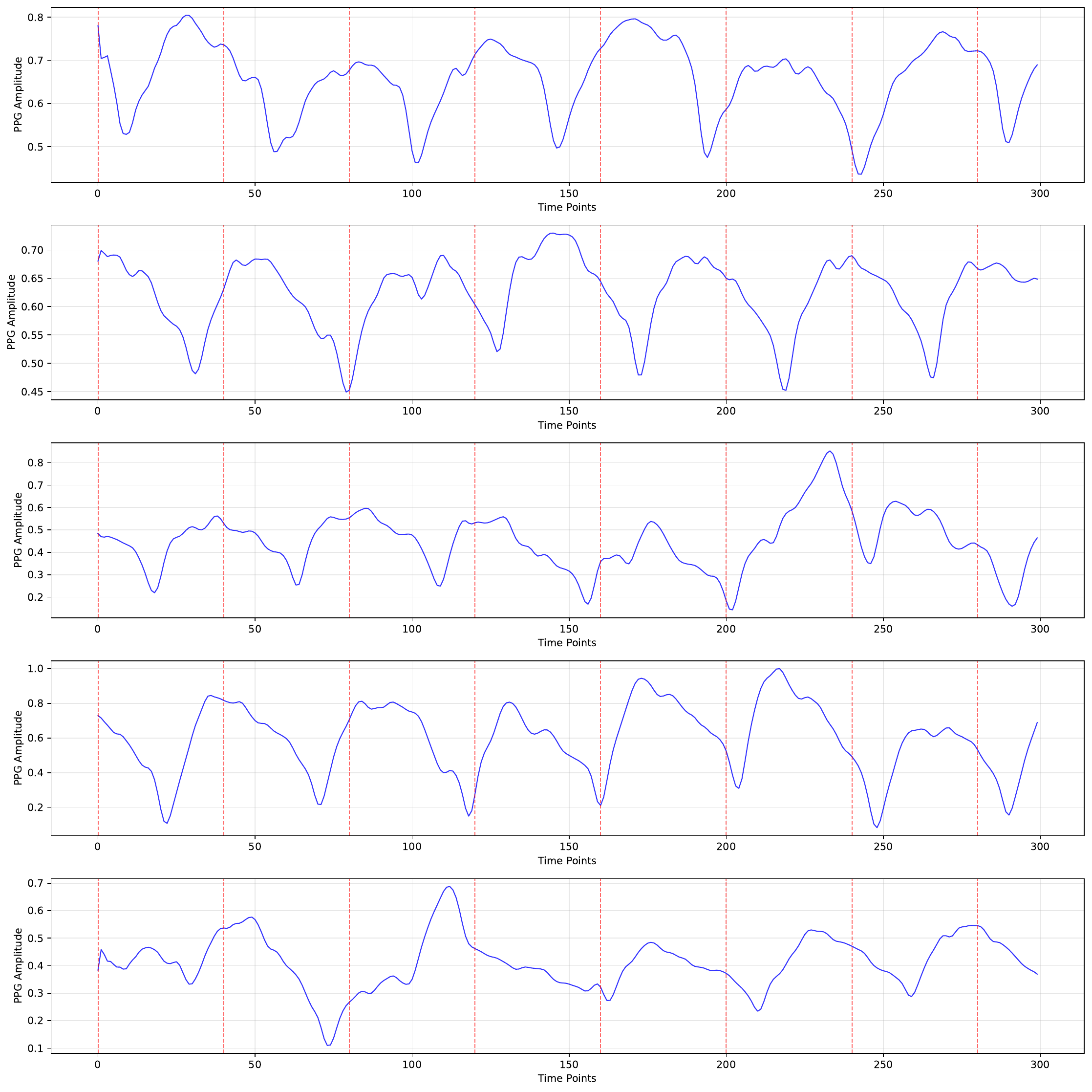}
        \vspace{-1em}
        \caption{Real PPG signals from the StanfordAF dataset, segmented into patches by red lines (patch size = 40).}
        \label{fig:vis_patch}
    \end{minipage}
    \vspace{-1em}
\end{figure*}

Given its wide range of applications, it is crucial to develop models that can learn generalizable representations from PPG signals and perform reliably across multiple downstream tasks. Recent studies have therefore introduced foundation models tailored to PPG signals~\cite{lee2025foundation, pillai2025papagei,chen2025gpt, saha2025pulse, erturk2025beyond}. Although these models demonstrate strong performance, deploying them on edge devices such as wearables remains difficult due to constraints on inference speed and memory usage. A natural solution is to leverage knowledge distillation (\textit{KD})~\cite{hinton2015distilling,gou2021knowledge} to compress large teacher models into a smaller, more efficient student models (Figure~\ref{fig:motivation}). However, the primary challenge lies in knowledge preservation, since vanilla \textit{KD} techniques may fail to transfer the nuanced understanding of PPG's unique characteristics.
This raises a critical question: \textit{What specific structural and temporal knowledge is essential for a PPG model, and how can it be effectively distilled from a teacher to a student?}

Most existing distillation methods concentrate on aligning output predictions~\cite{hinton2015distilling} or intermediate feature~\cite{romero2015fitnetshintsdeepnets} between a teacher and a student, namely \textbf{\vanilla{}}. Such approaches risk overlooking the local structural information that is central to PPG. In particular, waveform morphology within short temporal windows (patches) and structural rhythm between patches are essential for capturing both cardiovascular events and autonomic dynamics, yet these fine-grained patterns can be lost when only global prediction- or feature-level alignment is enforced. Moreover, recent PPG foundation models~\cite{pillai2025papagei, chen2025gpt} already adopt a patch-based representation, which naturally encodes local dynamics but remains underutilized during distillation.
To address this gap, we introduce \textbf{\method{}}, a distillation framework that augments vanilla prediction- and feature-level transfer with two novel patch-level strategies: morphology distillation, which enforces discriminability among local segments, and rhythm distillation, which preserves structural dependencies across patches. By explicitly transferring both global knowledge and local morphology–rhythm patterns, \method{} equips the student with richer PPG-specific representations. This design enables compact models that maintain strong task performance and are practical for on-device deployment. Across diverse benchmarks, \method{} achieves up to $21.80\%$ higher accuracy while reducing inference latency by up to $7\times$ and memory footprint by up to $19\times$ compared to the teacher, advancing the deployment of foundation-level PPG models in wearables. We discuss the related work in Appendix~\ref{app:related_work}

\section{Methodology}
We first introduce key notations. 
For PPG signal anslysis, given an input PPG signal \(X \in \mathbb{R}^{L} \), where \( L \) represents the length of the PPG signal, the goal is to predict the value \( Y \in \mathbb{R}^{1} \) for regression and the class \( Y \in \mathbb{R}^{C} \)for classification, where $C$ is the number of classes. Below, we propose and discuss several approaches to distill knowledge from a teacher PPG foundation model to a student. We start by adapting two \vanilla{} methods: prediction-matching and feature-matching. Next, we motivate and introduce our proposed \method{}, with patch-level matching strategies to distill additional patch-level local morphology-aware and structural rhythm information to the student.

\subsection{\vanilla{}}

The student produces predictions \(\hat{Y}_s\) and internal features \(H_s\in \mathbb{R}^{D}\). The teacher produces predictions \(\hat{Y}_t\) and internal features \(H_t\in \mathbb{R}^{D}\). The objective of \vanilla{} is:
\begin{equation}\label{eq:kd_obj}
    \min\nolimits_{\theta_s} \mathcal{L}_{sup}(Y, \hat{Y}_s) + \mathcal{L}_{\textit{KD}}^Y(\hat{Y}_t, \hat{Y}_s) + \mathcal{L}_{\textit{KD}}^H(H_t, H_s),
\end{equation}
where \(\theta_s\) is the parameter of the student; \(\mathcal{L}_{sup}\) is the supervised loss (e.g., MAE for regression, cross-entropy for classification); \(\mathcal{L}_{\textit{KD}}^Y\) and \(\mathcal{L}_{\textit{KD}}^H\) are the distillation loss terms that encourage student model to learn knowledge from teacher on both \textbf{prediction level}~\cite{hinton2015distilling} and \textbf{feature level}~\cite{romero2015fitnetshintsdeepnets}. However, \vanilla{} only matches the signal-level feature (i.e., \(\mathcal{L}_{\textit{KD}}^H\)), making it less effective at preserving the local morphology within each PPG segment and the structural rhythm across segments (Figure~\ref{fig:vis_patch}).

\subsection{\method{}}
\begin{figure*}[t!]
    \centering
    \includegraphics[width=0.99\textwidth]{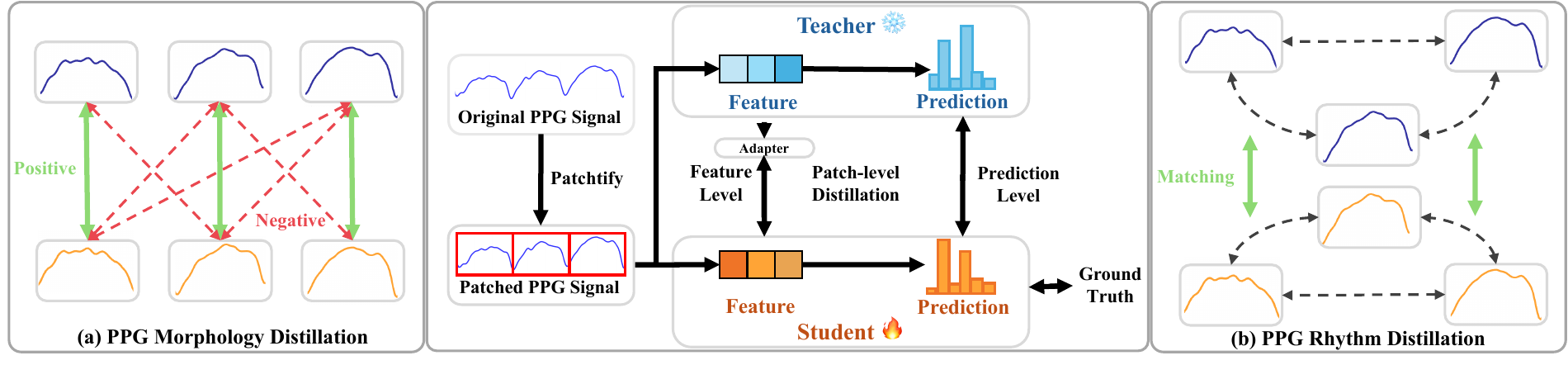}
    \vskip -1em
    \caption{Overall framework of \method{}.}
    \label{fig:method}
    \vskip -1em
\end{figure*}

In accordance with our intuition regarding preservation of local information of PPG signal, we propose a novel patch-level distillation framework, called~\method{} in Figure~\ref{fig:method}. Instead of focusing on matching global signal-level features, \method{} focuses on distilling knowledge about local morphology and rhythm by patch-level morphology and rhythm distillation. We note that the term \textit{morphology} here refers to data-driven local waveform representations within patches, rather than predefined or clinical morphological descriptors.

\textbf{Patchtify}, for most PPG foundation models~\cite {pillai2025papagei,chen2025gpt}, is the first step to process the original PPG signal $X$ to non-overlapping patches~\cite{nie2022time,liu2025cape, feng2025seizureformer}. Denote the patch length as $P$, then the patchifying process will generate a sequence of patches $X_p\in\mathbb{R}^{P\times N}$ where $N$ is the number of patches, $N = L/P$.

\paragraph{PPG Morphology Distillation}
Let the student and teacher produce features for a PPG patch sequence $X_p$ as
\(H_s^p\in\mathbb{R}^{N\times d_s}\) and \(H_t^p\in\mathbb{R}^{N\times d_t}\). Because \(d_s\) and \(d_t\) can differ, we
introduce a shared learnable linear adapter \(A\in\mathbb{R}^{d_t\times d_s}\) and form \(\tilde H_t^p=H_t^p A\). We then \(\ell_2\)-normalize patch
vectors row-wise, \(\hat H_{s/t}^p=\mathrm{norm}(H_{s/t}^p)\). We align the \(i\)-th student patch to the \(i\)-th
teacher patch and treat all other teacher patches as negatives.
The similarity matrix is
$Z=\frac{\hat H_s^p(\hat H_t^p)^\top}{\tau}\in\mathbb{R}^{N\times N}$,
where \(\tau\) is temperature. We use InfoNCE-style~\cite{oord2018representation} loss with one positive per row:
\[
\mathcal{L}_{mor}
=\frac{1}{N}\sum_{i=1}^{N}
\left(-\log\frac{\exp(Z_{ii})}{\sum_{j=1}^{N}\exp(Z_{ij})}\right).
\]
This objective encourages one-to-one alignment of local morphology across patches, allowing the student
to preserve the teacher’s patch-level morphology feature.

\paragraph{PPG Rhythm Distillation}
To keep the PPG rhythm (beat-to-beat periodicity and timing regularity), we transfer the teacher’s \emph{inter-patch relations} to the student rather than only aligning individual patch features. We form pairwise Euclidean distance matrices with normalization
$
[D_t]_{ij}=\big\| \phi(H_{t,i}^p)-\phi(H_{t,j}^p)\big\|_2,
[D_s]_{ij}=\big\| H_{s,i}^p-H_{s,j}^p\big\|_2,
$
The relational distillation loss matches these normalized structures with a smooth L1 penalty~\cite{park2019relational}:
\begin{equation}
\label{eq:rel}
\mathcal{L}_{rhy}
=\frac{1}{N(N-1)}\sum_{i\neq j}
\operatorname{smoothL1}\!\left( [\tilde D_s]_{ij},\ [\tilde D_t]_{ij}\right).
\end{equation}
This term penalizes discrepancies in relative inter-patch distances, thereby transferring the teacher’s structural knowledge of rhythm to the student.

\paragraph{Joint Optimization} \label{sec:joint_optimization}While training~\method{}, we jointly optimize both the PPG morphology and rhythm distillation losses in addition to the \vanilla{} losses. Therefore, the overall training loss that~\method{} adopts for the student is
$\mathcal{L} = \mathcal{L}_{sup} + \alpha \mathcal{L}_{\textit{KD}}^{Y} + \beta \mathcal{L}_{\textit{KD}}^{H} + \gamma (\mathcal{L}_{mor} + \mathcal{L}_{rhy})$, where $\alpha$, $\beta$, and $\gamma$ are hyper-parameters which mediate the strengths of each loss term.

\section{Experiment}
\paragraph{Experimental Setting}
To evaluate the effectiveness of \method{}, we benchmark it on both regression and classification tasks in PPG analysis, following GPT-PPG~\cite{chen2025gpt}. For regression, we use the DaLiA dataset~\cite{reiss2019deep}, where the model is required to estimate patients’ heart rates from PPG signals. For classification, we use the StanfordAF dataset~\cite{torres2020multi}, which targets atrial fibrillation (AF) detection. We adopt two PPG foundation models, GPT-PPG-19m~\cite{chen2025gpt} and PaPaGei~\cite{pillai2025papagei}, as teachers, and consider MLP as well as the lightweight GPT-PPG-1m variant of GPT-PPG as students. For regression, we report mean squared error (MSE) and mean absolute error (MAE)~\cite{ni2025we}. For classification, we report accuracy (Acc.) and F1 score. Further implementation details are provided in Appendix~\ref{app:implementation_details}.

\begin{table}[t!]
\centering
\caption{Performance comparison on DaLiA and StanfordAF. “+xx\%” values indicate the relative improvement in student performance after distillation.
}
\label{tab:main}
\resizebox{0.85\textwidth}{!}{%
\begin{tabular}{llcccc}
\toprule
\multicolumn{2}{c}{\textbf{Teacher Models}} & \multicolumn{2}{c}{\textbf{GPT-PPG-19m}~\cite{chen2025gpt}} & \multicolumn{2}{c}{\textbf{PaPaGei}~\cite{pillai2025papagei}} \\
\multicolumn{2}{c}{Metric} & MSE ($\downarrow$) & MAE ($\downarrow$) & MSE ($\downarrow$) & MAE ($\downarrow$) \\
\midrule
\multirow{7}{*}{DaLiA} 
 & Teacher       & 221.78 & 8.82 & 160.39 & 6.81 \\
 \cmidrule(lr){2-6}
 & MLP           & 581.77 & 17.87 & 581.77 & 17.87 \\
 & +\vanilla{} & 230.59{\scriptsize +60.36\%} & 10.74{\scriptsize +39.89\%} & 575.40{\scriptsize +1.10\%} & 17.84{\scriptsize +0.14\%} \\
 \cmidrule(lr){2-6}
 & GPT-PPG-1m~\cite{chen2025gpt}        & 255.07 & 10.08 & 255.07 & 10.08 \\
 & +\vanilla{} & 234.16{\scriptsize +8.20\%} & 9.44{\scriptsize +6.37\%} & 220.26{\scriptsize +13.65\%} & 8.38{\scriptsize +16.89\%} \\
 & \textbf{+\method{}}   & \textbf{215.36{\scriptsize +15.57\%}} & \textbf{8.34{\scriptsize +17.32\%}} & \textbf{202.31{\scriptsize +20.68\%}} & \textbf{7.90{\scriptsize +21.62\%}} \\
\midrule
\multicolumn{2}{c}{Metric} & Acc. ($\uparrow$) & F1 ($\uparrow$) & Acc. ($\uparrow$) & F1 ($\uparrow$) \\
\midrule
\multirow{7}{*}{StanfordAF} 
 & Teacher       & 0.93 & 0.88 & 0.83 & 0.70 \\
 \cmidrule(lr){2-6}
 & MLP           & 0.76 & 0.42 & 0.76 & 0.42 \\
 & +\vanilla{}   & 0.76{\scriptsize -0.09\%} & 0.54{\scriptsize +29.17\%} & 0.73{\scriptsize -4.31\%} & 0.41{\scriptsize -1.15\%} \\
 \cmidrule(lr){2-6}
 & GPT-PPG-1m~\cite{chen2025gpt}        & 0.81 & 0.64 & 0.81 & 0.64 \\
 & +\vanilla{} & 0.82{\scriptsize +0.80\%} & 0.65{\scriptsize +2.73\%} & 0.83{\scriptsize +1.83\%} & 0.67{\scriptsize +5.69\%} \\
 & \textbf{+\method{}}   & \textbf{0.87{\scriptsize +6.73\%}} & \textbf{0.77{\scriptsize +21.80\%}} & \textbf{0.88{\scriptsize +7.68\%}} & \textbf{0.77{\scriptsize +21.35\%}} \\
\bottomrule
\end{tabular}%
}
\vspace{-1em}
\end{table}

\paragraph{Results}

Table~\ref{tab:main} reports the effectiveness of the proposed \method{} compared with \vanilla{} on GPT-PPG-1m~\cite{chen2025gpt}. Since MLP does not patchify PPG signals, only \vanilla{} can be applied to it. Several key observations can be drawn from the results.   
\textbf{First}, \method{} consistently improves the performance of GPT-PPG-1m across both regression (DaLiA) and classification (StanfordAF) tasks. In particular, \method{} achieves up to a \textbf{+21.8\% relative F1 improvement} on StanfordAF and a \textbf{+13.7\% relative MSE improvement} on DaLiA, highlighting its strong and consistent gains across tasks. Notably, on the DaLiA dataset with GPT-PPG-19m as the teacher, GPT-PPG-1m trained with \method{} even outperforms its teacher while using $19\times$ fewer parameters, demonstrating that structural \textit{KD} can close, and even invert, the capacity gap between teacher and student.  
\textbf{Second}, MLP, even with \vanilla{}, fails to surpass GPT-PPG-1m, highlighting the limitation of its shallow architecture in modeling complex PPG dynamics.  
\textbf{Third}, \method{} consistently yields stronger performance than \vanilla{} when applied to GPT-PPG-1m, confirming that \method{} is more effective than \vanilla{}, particularly in transferring fine-grained rhythm and morphological cues that are crucial for PPG signal analysis. 
\textbf{Fourth}, on the DaLiA dataset, stronger teachers (e.g., PaPaGei) generally lead to better students, suggesting that high-quality teacher representations provide richer relational structure for distillation. However, this trend does not hold for the StanfordAF dataset, where the performance gap between teachers is smaller, and dataset-specific factors likely play a larger role. We conduct an ablation study and hyperparameter sensitivity analysis in Appendix~\ref{app:ablation}.

\begin{wrapfigure}{r}{0.5\textwidth}
    \vspace{-2.5em}
    \centering
    \includegraphics[width=0.85\linewidth]{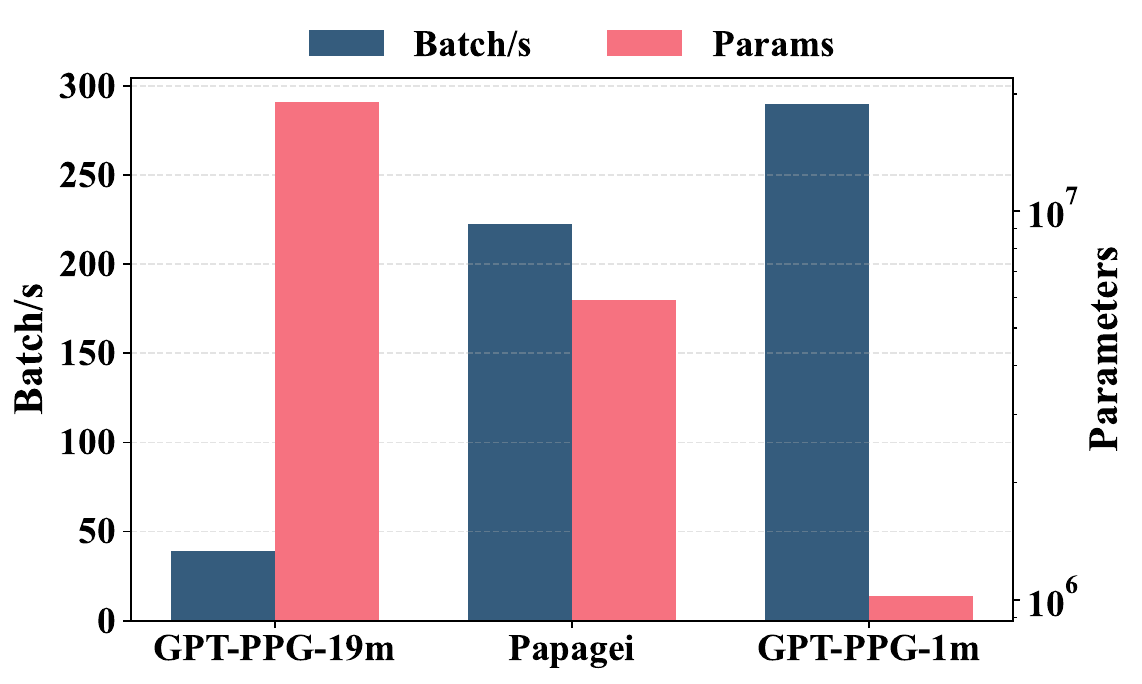}
    \vspace{-1em}
    \caption{Inference throughput (Batch/s) and parameter size comparison across GPT-PPG-19m, PaPaGei, and GPT-PPG-1m.}
    \label{fig:efficiency}
    \vspace{-2em}
\end{wrapfigure}

\paragraph{Efficiency Analysis}
To further evaluate the efficiency of \method{}, we compare throughput (measured in Batch/s) and model size (measured in number of parameters) across different models, as shown in Figure~\ref{fig:efficiency}.The results highlight two points.
First, foundation models such as GPT-PPG-19m and PaPaGei provide strong accuracy but suffer from low throughput and high memory cost, making them unsuitable for wearables.
Second, GPT-PPG-1m distilled with \method{} achieves the highest throughput with nearly $19\times$ fewer parameters, showing that compact students can retain strong performance while enabling efficient on-device inference. We provide detailed efficiency results in Appendix~\ref{app:efficiency}.

\section{Conclusion and Future Work}
We proposed \method{}, a distillation framework that combines prediction-, feature-, and patch-level strategies to transfer both global and local knowledge from large PPG foundation models to lightweight students. Experiments on heart rate estimation and atrial fibrillation detection show notable performance gains with much higher efficiency, enhancing the feasibility of real-world deployment of these models. Future work includes extending to more tasks and datasets, deeper analysis of the framework, and exploring diverse teacher models beyond foundation models.

\newpage
\bibliographystyle{plain}
\bibliography{neurips}


\newpage
\appendix
\section{Related Work}
\label{app:related_work}
\subsection{PPG Signal Analysis}
PPG has been used to estimate key physiological metrics, including heart rate~\cite{reiss2019deep, schmidt2018introducing}, heart rate variability~\cite{sarhaddi2022comprehensive}, blood glucose~\cite{ali2024comparison}, respiration rate~\cite{pimentel2016toward}, and blood pressure~\cite{wang2023pulsedb, he2022new}. Beyond general monitoring, PPG contributes to diagnostic applications by supporting the detection of cardiovascular conditions such as atrial fibrillation~\cite{torres2020multi, poh2018diagnostic}, reducing false arrhythmia alarms~\cite{clifford2015physionet}, and identifying hypoxia~\cite{lazazzera2020detection}. In addition, it is increasingly applied in mental health and wellness contexts, where it enables tracking of stress~\cite{zhu2023stress}, emotion~\cite{kontaxis2020photoplethysmographic}, and cognitive states such as focus~\cite{wang2024classifying}.

\subsection{Foundation Model for PPG Signal}

A foundation model is a large pre-trained model that learns general representations transferable to many downstream tasks~\cite{liu2025can}. 
Recent advances in foundation models for PPG signals can be categorized by their pre-training data sources. 
\textbf{Clinical or lab PPG-based models} include \textit{PaPaGei}~\cite{pillai2025papagei}, which leverages morphology-aware contrastive learning on 57,000 hours of clinical PPG and provides open-source weights, 
\textit{SiamQuality}~\cite{ding2024siamquality}, which enforces robustness to signal quality variations using over 36 million clinical PPG pairs, 
and \textit{GPT-PPG}~\cite{chen2025gpt}, which adapts generative transformers to ICU-collected PPG and demonstrates both predictive and denoising capabilities. 
In addition, \textit{REGLE}~\cite{yun2024unsupervised} employs autoencoders to extract disentangled embeddings from biobank-scale clinical PPG for genomic discovery and disease risk prediction, 
while \textit{TS2TC}~\cite{zhang2024general} introduces a generative self-supervised framework trained on the VitalDB dataset of surgical patients, aiming at physiological parameter estimation. 
\textbf{Field PPG-based models} directly address wearable applicability: \textit{Apple-PPG}~\cite{abbaspourazad2024largescale} is trained on data from more than 140K Apple Watch users and achieves strong generalization, though it remains closed-source, 
while \textit{Pulse-PPG}~\cite{saha2025pulse} represents the first open-source foundation model trained exclusively on large-scale wearable field PPG, showing improved robustness to motion noise and free-living conditions.

\subsection{Knowledge Distillation}
Knowledge distillation (\textit{KD})~\cite{hinton2015distilling} transfers knowledge from a larger, more complex model (teacher) to a smaller, simpler model (student) while maintaining comparable performance. 
By aligning the output distributions of teacher and student models, \textit{KD} provides richer training signals than hard labels alone, enabling the student to capture subtle patterns that the teacher has learned. In the context of time series signal, 
CAKD~\cite{xu2022contrastive} uses adversarial and contrastive learning for feature distillation without a specific design for time series, while LightTS~\cite{campos2023lightts} designs a \textit{KD} framework for ensemble classifiers, limiting its generality. Unlike these, TimeDistill~\cite{ni2025timedistill}
targets time series-specific patterns, such as multi-scale and
multi-period, pioneering cross-architecture \textit{KD} for time series analysis. To the best of our knowledge, we are the first attempt to apply the \textit{KD} technique to the PPG signal.

\section{Implementation Details} 
\label{app:implementation_details}
All experiments are implemented in PyTorch~\cite{paszke2019pytorch} and conducted on one NVIDIA L40S GPU. The teacher models are trained using their default configurations as reported in their respective papers. When using \method{} for distillation, the teacher model is frozen, and only the student is trained. 
Following GPT-PPG~\cite{chen2025gpt}, we set the patch size to 40.
We use Adam~\cite{kingma2014adam} for optimization. The initial learning rate is set by lr\_init=1e-5, and further adjustments are handled by the scheduler. A warmup and cosine annealing strategy is applied at the batch level with lr\_max=1e-3, eta\_min=1e-6, warm up ratio=25\%.
We apply early stopping with a patience value of 20 epochs. The batch size is set to 64. The temperature $\tau$ for patch-level contrastive distillation is set to $\tau=2$. We perform a hyperparameter search for $\alpha$, $\beta$ and $\gamma$ within the range \{0.1, 0.5\}. 

\begin{figure}[t] \centering
\includegraphics[width=0.45\textwidth]{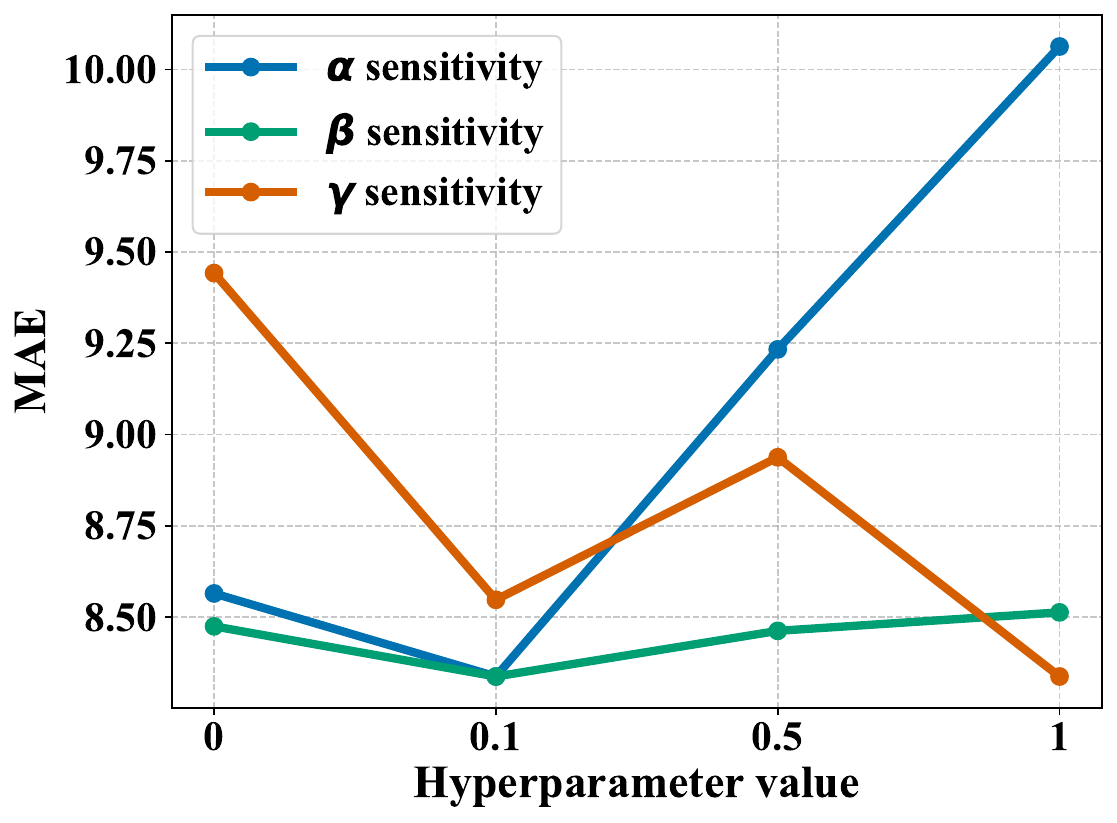}
    \caption{Effect of hyperparameters ($\alpha$, $\beta$, $\gamma$) on MAE for the DaLia dataset (Teacher: GPT-PPG-19m, Student: GPT-PPG-1m). }
    \label{fig:sensitivity}
\end{figure}

\section{Ablation study and Hyperparameter sensitivity} 
\label{app:ablation}
We varied $\alpha$, $\beta$, and $\gamma$ in the joint objective $\mathcal{L}$ in Section~\ref{sec:joint_optimization} to examine the effect of each loss term. As shown in Figure~\ref{fig:sensitivity}, $\alpha$ strongly influences performance: small values improve learning while large values degrade it. $\beta$ remains stable across settings, indicating feature-level distillation is less sensitive. $\gamma$ shows a non-monotonic trend, with $\gamma=1$ achieving the best MAE, confirming the importance of patch-level objectives for capturing morphology and rhythm.

\section{Full Results of Efficiency}
\label{app:efficiency}
\begin{table}[htbp]
\centering
\caption{Comparison on DaLiA dataset.}
\label{tab:dalia_efficiency}
\resizebox{0.75\textwidth}{!}{%
\begin{tabular}{lcccc}
\toprule
\textbf{DaLiA} & \textbf{GPT-PPG-19m} & \textbf{Papagei} & \textbf{MLP} & \textbf{GPT-PPG-1m} \\
\midrule
MAE     & 8.82 & 6.81 & 10.74 & 7.90 \\
Batch/s & 128.06 & 225.80 & 4248.70 & 291.50 \\
Params  & 19,018,417 & 5,917,197 & 41,473 & 1,017,197 \\
Memory (MB)  & 72.6 & 22.6 & 0.16 & 3.9 \\
\bottomrule
\end{tabular}%
}
\end{table}

\begin{table}[htbp]
\centering
\caption{Comparison on StanfordAF dataset.}
\label{tab:stanfordaf_efficiency}
\resizebox{0.75\textwidth}{!}{%
\begin{tabular}{lcccc}
\toprule
\textbf{StanfordAF} & \textbf{GPT-PPG-19m} & \textbf{Papagei} & \textbf{MLP} & \textbf{GPT-PPG-1m} \\
\midrule
F1       & 0.88 & 0.70 & 0.54 & 0.77 \\
Batch/s  & 39.19 & 222.30 & 1546.70 & 290.00 \\
Params   & 19,034,290 & 5,917,454 & 154,242 & 1,021,690 \\
Memory (MB) & 72.7 & 22.6 & 0.59 & 3.9 \\
\bottomrule
\end{tabular}%
}
\end{table}
Tables~\ref{tab:dalia_efficiency} and~\ref{tab:stanfordaf_efficiency} compare accuracy, inference throughput, and parameter efficiency across different models on the DaLiA and StanfordAF datasets. Several observations can be made. \textbf{First}, large foundation models such as GPT-PPG-19m achieve strong accuracy (MAE of 8.82 on DaLiA, F1 of 0.88 on StanfordAF) but come with high computational cost, processing fewer than 130 batches/s on DaLiA and fewer than 40 batches/s on StanfordAF. \textbf{Second}, PaPaGei provides a favorable trade-off, reducing parameters by about $3\times$ while maintaining competitive accuracy and substantially increasing throughput. \textbf{Third}, MLP achieves extremely high throughput (over 4000 batches/s on DaLiA), but its limited capacity results in a clear accuracy drop (MAE 10.74 on DaLiA, F1 0.54 on StanfordAF). \textbf{Finally}, GPT-PPG-1m, when distilled with \method{}, offers the best balance: it achieves accuracy close to or surpassing its teachers with only around 1M parameters, while running an order of magnitude faster than GPT-PPG-19m. These results highlight that \method{} enables lightweight models to approach the accuracy of large PPG foundation models while retaining significantly higher efficiency.

\end{document}